\documentclass[sigconf]{acmart}
\AtBeginDocument{%
  }

\setcopyright{acmlicensed}
\copyrightyear{2018}
\acmYear{2018}
\acmDOI{XXXXXXX.XXXXXXX}
\acmConference[Conference acronym 'XX]{Make sure to enter the correct
  conference title from your rights confirmation email}{June 03--05,
  2018}{Woodstock, NY}
\acmISBN{978-1-4503-XXXX-X/2018/06}

\usepackage{booktabs}
\usepackage{graphicx}
\usepackage{algorithm}
\usepackage{algorithmic}
\usepackage[most]{tcolorbox}
\usepackage{graphicx}
\usepackage{subcaption}
\usepackage{multirow}

\tcbuselibrary{listings, breakable}

\newtcbinputlisting[auto counter]{\promptbox}[3][]{%
    listing file={#3},
    listing only,
    breakable,
    enhanced,
    colback=blue!5!white,
    colframe=blue!75!black,
    fonttitle=\bfseries,
    title=Prompt \thetcbcounter: #2,
    attach title to upper=\par,
    after title={\hfill},
    #1
}

\newtcblisting{promptenv}[2][]{
    listing only,
    breakable,
    enhanced,
    colback=blue!5!white,
    colframe=blue!75!black,
    fonttitle=\bfseries,
    title=#2,
    listing options={
        basicstyle=\ttfamily\footnotesize,
        breaklines=true,
    },
    #1
}

\begin{document}

\title{Landscape-aware Automated Algorithm Design:\\ An Efficient Framework for Real-world Optimization}

\author{Haoran Yin}
\email{h.yin@liacs.leidenuniv.nl}
\orcid{0009-0005-7419-7488}
\affiliation{%
  \institution{LIACS, Leiden University}
  \city{Leiden}
  \country{Netherlands}
}

\author{Shuaiqun Pan}
\email{s.pan@liacs.leidenuniv.nl}
\orcid{0000-0001-7039-4875}
\affiliation{%
  \institution{LIACS, Leiden University}
  \city{Leiden}
  \country{Netherlands}
}

\author{Zhao Wei}
\email{wei_zhao@a-star.edu.sg}
\orcid{0000-0002-8173-9733}
\affiliation{%
  \institution{Agency for Science, Technology and Research (A*STAR)}
  \streetaddress{2 Fusionopolis Way}
  \country{Singapore}
}

\author{Jian Cheng Wong}
\email{wongj@a-star.edu.sg}
\orcid{0000-0002-3215-1888}
\affiliation{%
  \institution{Agency for Science, Technology and Research (A*STAR)}
  \streetaddress{2 Fusionopolis Way}
  \country{Singapore}
}

\author{Yew-Soon Ong}
\email{ong_yew_soon@a-star.edu.sg}
\orcid{0000-0002-4480-169X}
\affiliation{%
  \institution{Agency for Science, Technology and Research (A*STAR)}
  \streetaddress{2 Fusionopolis Way}
  \country{Singapore}
}

\author{Anna V. Kononova}
\email{a.kononova@liacs.leidenuniv.nl}
\orcid{0000-0002-4138-7024}
\affiliation{%
  \institution{LIACS, Leiden University}
  \streetaddress{Einsteinweg 55}
  \city{Leiden}
  \country{Netherlands}
}

\author{Thomas B{\"a}ck}
\email{t.h.w.baeck@liacs.leidenuniv.nl}
\orcid{0000-0001-6768-1478}
\affiliation{%
  \institution{LIACS, Leiden University}
  \streetaddress{Einsteinweg 55}
  \city{Leiden}
  \country{Netherlands}
}

\author{Niki van Stein}
\email{n.van.stein@liacs.leidenuniv.nl}
\orcid{0000-0002-0013-7969}
\affiliation{%
  \institution{LIACS, Leiden University}
  \streetaddress{Einsteinweg 55}
  \city{Leiden}
  \country{Netherlands}
}

\renewcommand{\shortauthors}{Trovato et al.}

\begin{abstract}
The advent of Large Language Models (LLMs) has opened new frontiers in automated algorithm design, giving rise to numerous powerful methods. However, these approaches retain critical limitations: they require extensive evaluation of the target problem to guide the search process, making them impractical for real-world optimization tasks, where each evaluation consumes substantial computational resources. This research proposes an innovative and efficient framework that decouples algorithm discovery from high-cost evaluation. Our core innovation lies in combining a Genetic Programming (GP) function generator with an LLM-driven evolutionary algorithm designer. The evolutionary direction of the GP-based function generator is guided by the similarity between the landscape characteristics of generated proxy functions and those of real-world problems, ensuring that algorithms discovered via proxy functions exhibit comparable performance on real-world problems. Our method enables deep exploration of the algorithmic space before final validation while avoiding costly real-world evaluations. We validated the framework's efficacy across multiple real-world problems, demonstrating its ability to discover high-performance algorithms while substantially reducing expensive evaluations. This approach shows a path to apply LLM-based automated algorithm design to computationally intensive real-world optimization challenges.
\end{abstract}

\ccsdesc[300]{Theory of computation~Optimization with randomized search heuristics}
\ccsdesc[500]{Computing methodologies~Heuristic function construction}
\ccsdesc[100]{Computing methodologies~Natural language generation}

\keywords{Program synthesis, Simulation optimization, Genetic programming, Large language models, Landscape analysis}
\maketitle
\section{Introduction}
Optimization problems are prevalent in real-world domains such as engineering, manufacturing, scientific simulation, and product design~\cite{vanderplaats1996design, DBLP:books/daglib/0034477, DBLP:journals/nca/Abualigah0KAIAM22, alemao2021smart}. These problems typically exhibit characteristics including high dimensionality, multimodality, noise, and expensive computation, which pose significant challenges to traditional optimization methods~\cite{jin2005evolutionary,forrester2008engineering}. With the continuous advancement of computational intelligence, automated algorithm design (AAD) has emerged as a crucial approach to deal with complex optimization problems~\cite{hutter2019automated}. Its objective is to generate or configure high-performance optimization algorithms using automated methods, thus reducing the work of manual design and improving the efficiency of problem-solving.

In recent years, AAD methods based on Large Language Models (LLMs) have emerged, demonstrating formidable capabilities in algorithm synthesis and code generation~\cite{liu2024evolution,van2024llamea}. For instance, frameworks such as the Large Language Model Evolutionary Algorithm (LLaMEA) can now automatically generate competitive meta-heuristic algorithms through natural language interaction and iterative optimization using Evolution Strategy (ES)~\cite{van2024llamea,rechenberg1978evolutionsstrategien}. Nevertheless, these approaches face a fundamental bottleneck: substantial evaluations of the target problem are typically required to guide algorithm discovery~\cite{yang2023large,yin2025optimizing}. In real-world settings, each evaluation can take minutes to days, so the resulting total evaluation time (or budget) can be prohibitive, making these methods difficult to implement in practice.

To reduce reliance on such expensive evaluations, landscape analysis offers powerful tools to understand the structural characteristics of optimization problems. By analyzing landscape features, attributes such as modality, ruggedness, and neutrality can be characterized, providing a basis for predicting algorithmic performance~\cite{mersmann2011exploratory,kerschke2019automated}. Existing studies have shown that Genetic Programming (GP) can automatically generate mathematically complex expressions, allowing the construction of proxy functions for real-world problems that approximate key characteristics of the target landscape~\cite{koza1994genetic,tian2020recommender,DBLP:conf/ijcci/LongVKKYBS23,long2024generating}.

\begin{figure}
    \centering
    \includegraphics[width=0.95\linewidth]{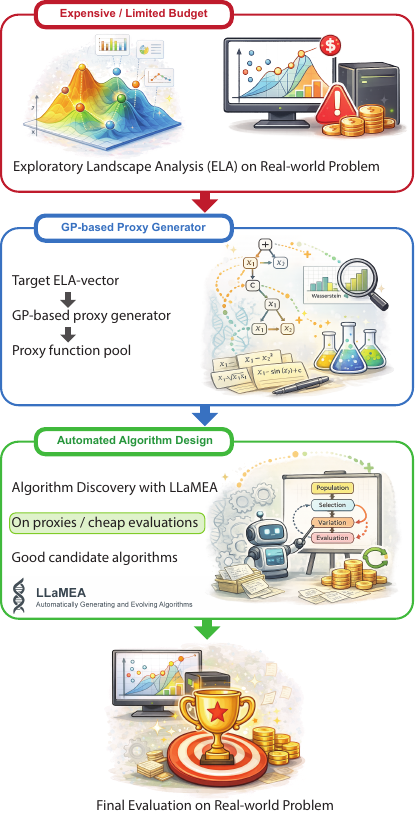}
    \caption{An overview of the proposed landscape-aware AAD framework. It first generates low-cost proxy functions using GP to match the landscape of the real problem, then uses LLMs to discover algorithms on these proxies, and finally validates only top candidates on the real world problem.}
    \label{fig:framework}
\end{figure}

To address the high computational cost of applying AAD methods to real-world single-objective continuous problems, this paper proposes a novel and highly efficient landscape-aware framework for AAD, which is highlighted in Figure~\ref{fig:framework}. This framework jointly leverages a GP-based proxy function generator and an LLM-based algorithm designer to decouple algorithm discovery from expensive real-world evaluations. Specifically, we first use GP to generate a set of proxy functions that approximate key characteristics of the target landscape. By minimizing the Wasserstein distance between the empirical distributions of the landscape features of the proxy and the real-world problem~\cite{kantorovich1960mathematical}, we encourage the proxy functions to match the landscape characteristics of the target problem. Subsequently, the LLM conducts algorithm discovery by evaluating candidate algorithms on proxy functions, drastically reducing the real-world evaluation budget. Only a small number of promising candidates are finally validated on the real problem, resulting in high-performance algorithms for practical optimization tasks.


Our main contributions can be summarized as follows:
\begin{itemize}
    \item We propose a novel landscape-aware framework that effectively decouples algorithm discovery from costly real-world evaluations, enabling efficient exploration of the algorithmic space.
    \item We integrate a landscape-similarity-guided proxy mechanism into the LLamEA framework, bridging classical landscape theory with LLM-driven automated algorithm design. By minimizing the Wasserstein distance between GP-generated functions and LLM-specific search landscapes, we create a proxy that allows LLamEA to explore the algorithmic space.
    \item We demonstrate a practical and resource-efficient verification strategy, where only top-performing candidates undergo final real-world validation, allowing high-performance algorithm discovery under strictly constrained evaluation budgets.
\end{itemize}

\section{Related Work}
\label{sec:related_work}
To position our work within the broader literature, this section reviews three closely related areas: landscape analysis for characterizing optimization problems, GP-based function evolution, and automated algorithm design.

\subsection{Landscape Analysis}
Landscape analysis characterizes continuous black-box optimization problems using measurable properties of the fitness landscape, and has long been used to link instance structure with algorithm behavior~\cite{DBLP:conf/icga/JonesF95,DBLP:journals/isci/MalanE13}. 
Common difficulty indicators include the fitness--distance correlation (FDC)~\cite{DBLP:conf/icga/JonesF95} and ruggedness estimates derived from autocorrelation along random walks~\cite{weinberger1990correlated}, while information-theoretic measures provide complementary signals of modality and smoothness~\cite{DBLP:journals/tec/MunozKH15}. 
Building on these ideas, Exploratory Landscape Analysis (ELA) defines a comprehensive feature set for continuous optimization that is widely used for feature-based benchmarking and performance modeling; toolkits such as \texttt{flacco} further enable standardized large-scale feature extraction~\cite{mersmann2011exploratory,kerschke2019comprehensive}. 
Landscape features are also central to algorithm selection and performance prediction under Rice’s framework~\cite{DBLP:journals/ac/Rice76,DBLP:journals/aim/Kotthoff14}, including automated approaches that learn mappings from ELA features to algorithm performance or rankings~\cite{kerschke2019automated}. 
Related instance-space analysis similarly embeds problem instances into a feature space to study how algorithm strengths vary across regions of that space~\cite{DBLP:journals/cor/Smith-MilesBWL14}.

\subsection{Function Evolution}
Evolving mathematical functions and executable structures is a central topic in GP and symbolic regression. 
Since Koza’s formulation, GP has been used to evolve tree-structured expressions without assuming a parametric model~\cite{koza1994genetic,DBLP:books/lulu/PoliLM2008}. 
Modularity is crucial for scaling expressivity: Automatically Defined Functions (ADFs) promote the reuse of evolved subroutines and support the construction of hierarchical programs~\cite{koza1994genetic}. 
Modern GP further emphasizes generalization and model simplicity through bloat control and parsimony mechanisms~\cite{DBLP:series/ncs/PoliM14}, alongside constrained representations that enforce syntactic or semantic validity, including grammar-based approaches and grammatical evolution~\cite{DBLP:books/daglib/0008988}, as well as strongly typed GP~\cite{DBLP:journals/ec/Montana95}. 
To reduce the cost of fitness evaluation, the surrogate-assisted GP replaces expensive evaluations with learned approximations when possible~\cite{DBLP:journals/ec/HildebrandtB15}. 
Most closely related to our goal, recent work uses landscape descriptors (e.g., ELA) to guide GP toward generating cheap proxy functions that resemble the characteristics of the target problem~\cite{DBLP:conf/ijcci/LongVKKYBS23}.
We extend this direction by generating proxies specifically to support transferable algorithm discovery under a limited budget of expensive target evaluations.


\subsection{Automated Algorithm Design}
AAD covers hyperparameter optimization, algorithm configuration, and meta-learning. Practical systems such as Sequential Model-Based Algorithm Configuration (\texttt{SMAC}) and Iterated Racing for Automatic Algorithm Configuration (\texttt{irace}) have shown strong results for tuning parameterized solvers~\cite{hutter2011sequential,lopez2016irace,kerschke2019automated}. 
Moving beyond parameter tuning, hyper-heuristics aim to automate the selection or construction of heuristics and search strategies, enabling higher-level control over the search process~\cite{burke2013hyper}. 
More recently, LLMs have been incorporated into heuristic-search pipelines to generate and iteratively improve algorithmic components, as demonstrated by FunSearch and LLaMEA~\cite{DBLP:journals/nature/RomeraParedesBNBKDREWFKF24,van2024llamea}. 
However, most of these approaches still depend on large numbers of task evaluations to provide feedback and selection pressure, which is impractical when each evaluation is costly. 
We address this limitation by separating discovery from expensive target evaluations: proxy functions are evolved under landscape-similarity constraints and then used to guide LLM-driven synthesis, with only a small budget reserved for final validation on the real tasks.

\section{Methodology}
\label{sec:methodology}
This section proposes an AAD framework based on landscape awareness, combining proxy function generation through GP with algorithm discovery driven by LLMs.
We first describe how to extract ELA features and employ them as similarity bridges, then detail the proxy function generation workflow, and finally explain how to adapt LLM-based designers to the proxy function-driven discovery context.

\begin{table}[!tb]
\caption{Definitions of numbers, variables, and operators for constructing proxy functions.}
\label{tab:operators}
\resizebox{\columnwidth}{!}{%
\begin{tabular}{@{}cccc@{}}
\toprule
Notation                         & Meaning                  & Type                     & Syntax                                                        \\ \midrule
\texttt{a}      & a real constant          & Number                   & $a$                                                           \\
\texttt{rand}   & a random number          & Number                   & $rand$                                                        \\
\texttt{index}  & index vector             & Decision Variable        & $(1, ..., d)$                                                 \\
\texttt{x}      & decision vector          & Decision Variable        & $(x_1, ..., x_d)$                                             \\
\texttt{add}    & addition                 & Binary Operator          & $x+a$                                                         \\
\texttt{sub}    & subtraction              & Binary Operator          & $x-a$                                                         \\
\texttt{mul}    & multiplication           & Binary Operator          & $a \dot x$                                                    \\
\texttt{div}    & division                 & Binary Operator          & $a/x$                                                         \\
\texttt{neg}    & negative                 & Unary Operator           & $-x$                                                          \\
\texttt{rec}    & reciprocal               & Unary Operator           & $1/x$                                                         \\
\texttt{multen} & multiplying by ten       & Unary Operator           & $10x$                                                         \\
\texttt{square} & square                   & Unary Operator           & $x^2$                                                         \\
\texttt{abs}    & absolute value           & Unary Operator           & $\left | x \right |$                                          \\
\texttt{sqrt}   & square root              & Unary Operator           & $\sqrt{\left | x \right | }$                                  \\
\texttt{exp}    & exponent                 & Unary Operator           & $e^x$                                                         \\
\texttt{ln}    & natural logarithm                & Unary Operator           & $\ln{\left | x \right | }$                                    \\
\texttt{sin}    & sine                     & Unary Operator           & $\sin{2\pi x}$                                                \\
\texttt{cos}    & cosine                   & Unary Operator           & $\cos{2\pi x}$                                                \\
\texttt{round}  & rounded value            & Unary Operator           & $\left \lceil x \right \rceil$                                \\
\texttt{sum}    & sum of vector            & Vector-Oriented Operator & $\sum_{i=1}^{d}x_i$                                           \\
\texttt{mean}   & mean of vector           & Vector-Oriented Operator & $\frac{1}{d} \sum_{i=1}^{d}x_i$                               \\
\texttt{cum}    & cumulative sum of vector & Vector-Oriented Operator & $\left ( \sum_{i=1}^{1}x_i, ...,  \sum_{i=1}^{d}x_i \right )$ \\
\texttt{prod}   & product of vector        & Vector-Oriented Operator & $\prod_{i=1}^{d}x_i$                                          \\
\texttt{max}    & maximum value of vector  & Vector-Oriented Operator & $\max_{i=1,..,d} x_i$                                         \\ \bottomrule
\end{tabular}%
}
\end{table}

\subsection{Landscape Characterization}
\label{Methodology:ELA}
We strategically select 7 classic sets
of ELA features that are low-cost and efficient to measure the landscape characteristics of single-objective continuous optimization problems based on current research~\cite{munoz2015effects,vskvorc2020understanding,long2023bbob}, including $y$\textit{-distribution}, \textit{level set}, \textit{meta model}, \textit{dispersion}, \textit{nearest better clusterin (NBC)}, \textit{principal component analysis (PCA)}, and \textit{Information Content of Fitness Sequences (ICoFiS)}~\cite{KerschkeT2019flacco}. To calculate the ELA feature, we sample $coef_{ELA} \times D$
points for the problem in dimension $D$, where $coef_{ELA}$ is the coefficient controlling the sample size, and randomly select $rate_{ELA}$ points from $coef_{ELA} \times D$ points to calculate the ELA feature and repeat it $n_{ELA}$ times to build the distributions of ELA features and avoid the impact of data fluctuations. We compute the correlation coefficients between the features and remove any retained feature with an absolute correlation coefficient that exceeds $threshold_{corr}$ compared to any other retained feature. This reduces redundancy and enhances the robustness of subsequent similarity metrics.

\subsection{Proxy Function Generator}
\label{Methodology:Proxy}

Based on current research on GP-based function generators~\cite{tian2020recommender,DBLP:conf/ijcci/LongVKKYBS23,long2024generating}, our proxy functions are defined by 2 types of numbers, 2 variables, and 20 operators, which are shown in Table \ref{tab:operators}. The proxy function generation system with which we are going to experiment is constructed on the basis of the Distributed Evolutionary Algorithms in Python (DEAP) framework~\cite{DEAP_JMLR2012}. This framework provides one-point crossover and subtree mutation operations for the implementation of the proxy function generator~\cite{syswerda1989uniform, koza1994genetic}. The Algorithm~\ref{alg:gp-fg} illustrates the system workflow, where the Half-and-Half initialization method randomly selects either the Growth rule or the Full rule when generating each initial individual~\cite{koza1994genetic}. The Growth approach allows random selection of terminal or non-terminal nodes before reaching maximum depth, yielding irregularly shaped trees; the Full approach selects only non-terminal nodes before reaching maximum depth, generating trees with uniform branch depths.

\begin{algorithm}[!tb]
\caption{Landscape-Guided Function Generation}
\label{alg:gp-fg}
\begin{algorithmic}[1]
\REQUIRE{$\mathbf{X}$: designed sampling points, $\mathbf{t}$: target ELA feature vector, $P$: populations, $n_{pop}$: population size, $n_{gen}$: number of generations, $p_c$: crossover probability, $p_m$: mutation probability, $min_{depth}$: minimum depth of tree structures, $max_{depth}$: maximum depth of tree structures}
\ENSURE{Best generated functions $f^*$}
\STATE Initialize population $P_0$ using half-and-half method, the depth is randomly decided between $min_{depth}$ and $max_{depth}$ for each individual $ind$.
\FOR{$g = 1$ to $n_{gen}$}
    \FOR{each individual $ind$ in $P_{g-1}$}
        \STATE Compile tree to function: $f \gets \text{compile}(ind)$
        \STATE Evaluate $f$ on $\mathbf{X}$: $\mathbf{y} \gets f(\mathbf{X})$
        \IF{$\mathbf{y}$ is invalid (NaN, Inf, or constant)}
            \STATE $fitness \gets \text{penalty}$
        \ELSE
            \STATE Compute ELA features: $\mathbf{c} \gets \text{ela\_features}(\mathbf{X}, \mathbf{y})$
            \STATE $fitness \gets \text{wasserstein\_distance}(\mathbf{c}, \mathbf{t})$
        \ENDIF
        \STATE $ind.fitness \gets fitness$
    \ENDFOR
    \STATE Select parents: $P' \gets \text{tournament\_selection}(P_{g-1})$
    \STATE Create offspring $P_g$ by applying:
        \STATE \quad - One-point crossover with rate $p_c$
        \STATE \quad - Subtree mutation with rate $p_m$
\ENDFOR
\STATE Extract best functions $f^*$
\RETURN $f^*$
\end{algorithmic}
\end{algorithm}



In terms of similarity metrics, we selected the Wasserstein distance as the measure of similarity between the ELA features of the proxy functions and those of real-world problems~\cite{vaserstein1969markov}.

\subsection{Proxy-Driven Algorithm Discovery}

Research on the use of LLMs for AAD is booming. Our experiments focus on the LLaMEA framework, which has been demonstrated to be both reliable and superior~\cite{van2024llamea}. LLM-based algorithm discovery, such as LLaMEA, requires extensive LLM conversations to iterate the algorithm. And certainly, each iteration necessitates substantial evaluation based on real-world problems, which is not industrial-friendly, as most real-world problems' evaluations are expensive.

The core innovation lies in introducing GP as an offline, low-cost knowledge distillation module that generates reusable high-quality proxy functions that are then used by the LLM to automatically design algorithms with a significantly reduced need for expensive evaluations.

In our framework, task prompt is slightly different from commonly used one, as we need to inform LLM that the generated algorithm is evaluated based on proxy functions, not benchmark problems, such as BBOB, any more.
\begin{promptenv}{Enhanced prompt with primitive guidance}
The optimization algorithm should handle a wide range of tasks, which is evaluated on the similar problems of a real-world problem. Your task is to write the optimization algorithm in Python code. The code should contain an `__init__(self, budget, dim)` function and the function `def __call__(self, func)`, which should optimize the black box function `func` using `self.budget` function evaluations.
The func() can only be called as many times as the budget allows, not more. Each of the optimization functions has a search space between func.bounds.lb (lower bound) and func.bounds.ub (upper bound). The dimensionality can be varied.
Give an excellent and novel heuristic algorithm to solve this task and also give it a one-line description with the main idea.
\end{promptenv}
\section{Real-world Problems}
\label{sec:problems}
To validate the framework's robustness, we conducted benchmarks on complex optimization tasks within meta-surface design and photonic optimizations. These domains are characterized by computationally prohibitive evaluations and intricate landscapes, providing a rigorous environment to test the efficacy of our framework under high-cost constraints. The following sections detail the problem applications, parameter configurations, and their specific landscape challenges.

\subsection{Meta-surface Design}
A meta-surface is a two-dimensional planar material composed of subwavelength structures capable of precisely manipulating light waves (such as phase, amplitude, and polarization) akin to conventional optical elements. Its objective is: given a desired optical function (for instance, achieving total internal reflection of light at a specific frequency or bending it at a particular angle), how can one automatically design the microstructural pattern required to realize that function? The instance we obtain was originally developed by Jiang et al. and has been studied by Dai et al. recently, and both the real solver and the surrogate model are available~\cite{jiang2018quasi,dai2022slmgan}. The task is to find a meta-surface structure whose physical properties closely match the target profile.

\subsection{Bragg Mirror Design}
Bragg mirrors (also termed Bragg reflectors) comprise two or more semiconducting or dielectric materials arranged in alternating layers, achieving exceptionally high reflectivity within specific optical bands~\cite{bragg1914mr}. This structure finds significant applications in numerous fields, such as the construction of acoustic reflectors and filters~\cite{priyadarshini2024distributed}, the analysis of the crystalline structure of materials~\cite{mihai2015metallic}, the improvement of the efficiency of solar cells~\cite{jiang2018design}, the monitoring of changes in physical quantities such as temperature and pressure~\cite{gryga2022distributed}, and the strengthening of light-matter interactions in quantum computing and quantum information fields~\cite{malak2014beyond}. We have two instances of this problem, called \textit{mini-Bragg} and \textit{Bragg}, which are from the testbed built by Bennet et al.~\cite{bennet:hal-02613161} Their settings can be found in Table~\ref{tab:instances}, where different columns, respectively, represent the parameters of the corresponding material. The goal is to maximize reflectivity at 600 nm wavelength.

\begin{table}[!tb]
\caption{The parameter settings for photonic instances. Different columns of thickness and permittivity correspond to different materials.}
\label{tab:instances}
\resizebox{\columnwidth}{!}{%
\begin{tabular}{@{}cccccccc@{}}
\toprule
instances         & \multicolumn{2}{c}{\textit{mini-Bragg}}                & \multicolumn{2}{c}{\textit{Bragg}}                     & \textit{ellipsometry} & \multicolumn{2}{c}{\textit{photovoltaic}}            \\ \midrule
layers            & \multicolumn{2}{c}{10}                        & \multicolumn{2}{c}{20}                        & 1            & \multicolumn{2}{c}{10}                      \\
materials         & \multicolumn{2}{c}{2}                         & \multicolumn{2}{c}{2}                         & 1            & \multicolumn{2}{c}{2}                       \\
min thickness(nm) & 0                     & 0                     & 0                     & 0                     & 50           & 30                   & 30                   \\
max thickness(nm) & 218                   & 218                   & 218                   & 218                   & 150          & 250                  & 250                  \\
min permittivity  & \multirow{2}{*}{1.96} & \multirow{2}{*}{3.24} & \multirow{2}{*}{1.96} & \multirow{2}{*}{3.24} & 1.1          & \multirow{2}{*}{2.0} & \multirow{2}{*}{3.0} \\
max permittivity  &                       &                       &                       &                       & 3.0          &                      &                      \\ \bottomrule
\end{tabular}%
}
\end{table}

\subsection{Ellipsometry Inverse Problem}
Ellipsometry is a non-destructive optical measurement technique capable of determining parameters such as film thickness, refractive index, and absorption coefficient~\cite{rothen1945ellipsometer}. In semiconductor fabrication, the properties of the films critically influence the performance of the circuit~\cite{zollner2013spectroscopic}. Solving the ellipsometry inverse problem improves the precision of the process control~\cite{toomey2024tackling}. This technique is also extensively applied in the fields of new energy and chemical engineering to investigate solar cell materials, nanostructured films, and chemical coating properties~\cite{fujiwara2018spectroscopic}. Solving the ellipsometry inverse problem not only advances data analysis methods in materials science but also enhances the industrial applicability of thin-film technologies~\cite{gonccalves2002fundamentals,schubert2004infrared,hinrichs2018ellipsometry}. The instance \textit{ellipsometry} is from the same testbed as \textit{mini-Bragg} and \textit{Bragg}. Its setting is shown in Table~\ref{tab:instances}.

\subsection{Photovoltaic Design}
This problem aims at finding the design of precision anti-reflective (AR) coatings for solar cells. Such coatings typically comprise dielectric layers with alternating refractive indices. By suppressing surface reflection losses, carefully engineered AR coatings can substantially enhance the photovoltaic conversion efficiency of solar cells~\cite{ji2022recent}. The challenge lies in achieving broadband, omnidirectional anti-reflection: single-layer coatings only eliminate reflection within narrow bands, necessitating multi-layer or gradient refractive index designs to reduce reflectance across the entire solar spectrum. Optimizing such coatings constitutes a complex inverse photonic design problem, with the performance space containing numerous local minima due to wave interference effects at multiple wavelengths~\cite{bennet2024illustrated}. Table~\ref{tab:instances} shows the setting for \textit{photovoltaic}, which is from the same testbed as \textit{mini-Bragg}, \textit{Bragg}, and \textit{ellipsometry}. The goal is to maximize absorption within the desired wavelength range.

\section{Experimental Setup}
\label{sec:setup}

In this section, we detail an experimental framework designed to evaluate the synergy between the GP-based proxy function generator and the LLM-based algorithm designer. Our primary investigation focuses on whether synthetic proxy functions can effectively guide algorithm discovery, thereby mitigating the prohibitive computational costs typically associated with direct real-world evaluations.

\subsection{Research Hypotheses}
We test the following three hypotheses:
\begin{itemize}
    \item H1 (Performance Efficacy): Algorithms discovered through landscape-aware proxy functions will demonstrate superior performance on real-world optimization tasks compared to those evolved using standardized artificial benchmarks, such as the BBOB suite.
    \item H2 (Computational Efficiency): The proposed framework significantly curtails the search overhead, requiring at least a tenfold reduction (one order of magnitude) in high-cost real-world evaluations compared to direct-optimization discovery methods.
    \item H3 (Competitive Parity and Superiority): We hypothesize that the algorithms discovered by our framework will demonstrate optimization capabilities that are commensurate with or superior to established meta-heuristics, specifically Differential Evolution (DE) and its state-of-the-art variant, LSHADE~\cite{,storn1997differential,aranguren2021improving}, when evaluated on diverse real-world instances.
\end{itemize}

\begin{figure*}[t]
    \centering
    \begin{subfigure}[b]{0.66\columnwidth}
        \centering
        \includegraphics[width=\textwidth]{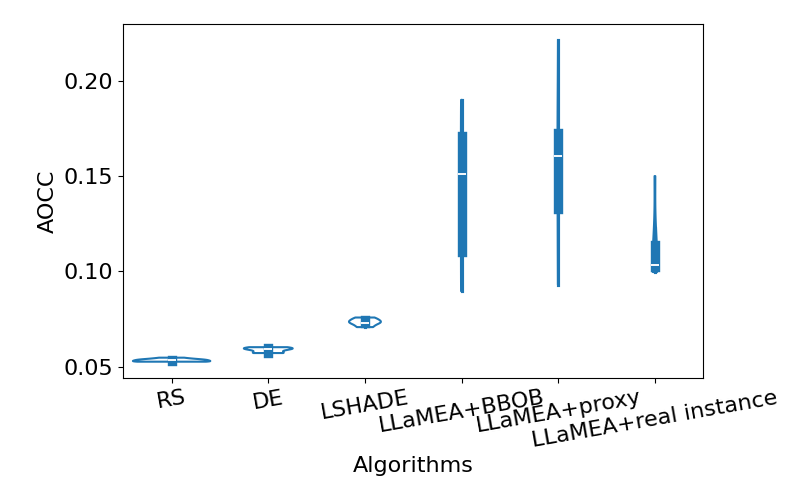}
        \caption{\textit{meta-surface} (surrogate model)}
        \label{fig:AOCC_meta_surface_surrogate}
    \end{subfigure}
    \begin{subfigure}[b]{0.66\columnwidth}
        \centering
        \includegraphics[width=\textwidth]{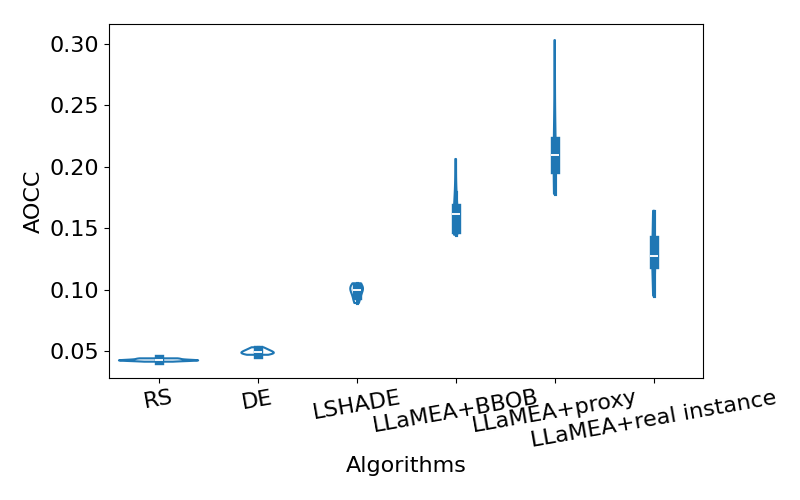}
        \caption{\textit{meta-surface} (real solver)}
        \label{fig:AOCC_meta_surface_solver}
    \end{subfigure}
    \begin{subfigure}[b]{0.66\columnwidth}
        \centering
        \includegraphics[width=\textwidth]{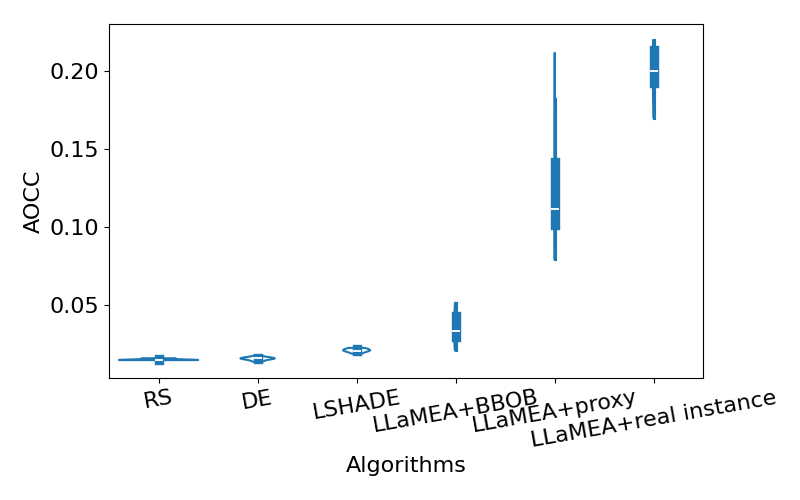}
        \caption{\textit{mini-Bragg}}
        \label{fig:AOCC_10bragg}
    \end{subfigure}
    \begin{subfigure}[b]{0.66\columnwidth}
        \centering
        \includegraphics[width=\textwidth]{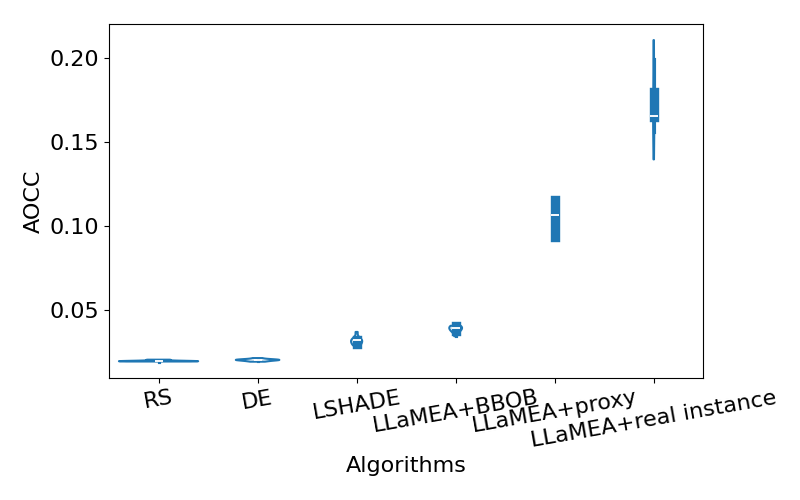}
        \caption{\textit{Bragg}}
        \label{fig:AOCC_20bragg}
    \end{subfigure}
    \begin{subfigure}[b]{0.66\columnwidth}
        \centering
        \includegraphics[width=\textwidth]{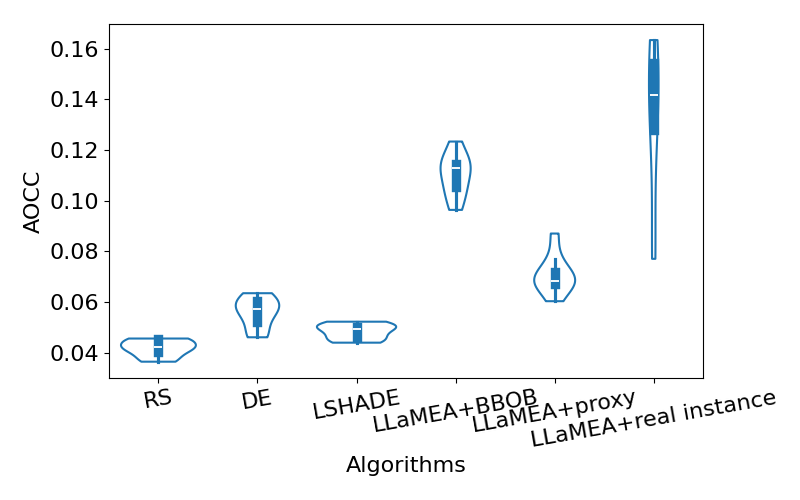}
        \caption{\textit{ellipsometry}}
        \label{fig:AOCC_ellipsometry}
    \end{subfigure}
    \begin{subfigure}[b]{0.66\columnwidth}
        \centering
        \includegraphics[width=\textwidth]{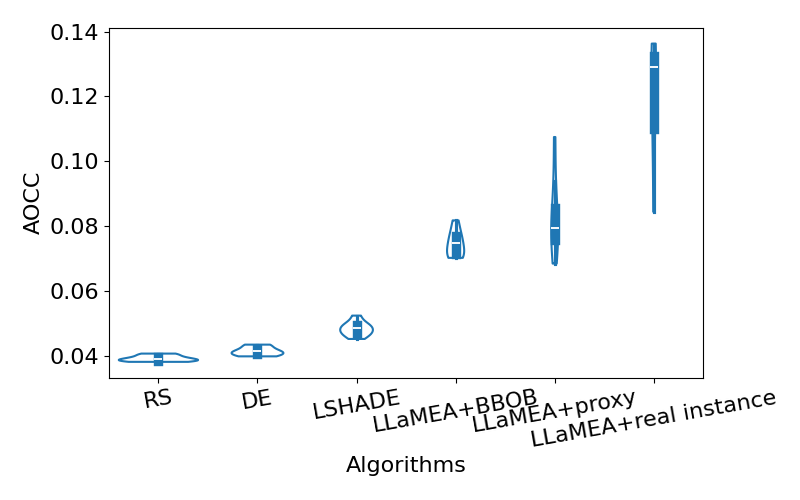}
        \caption{\textit{photovoltaic}}
        \label{fig:AOCC_photovoltaic}
    \end{subfigure}
    \caption{The distribution of AOCC values across 10 runs of different algorithms on real-world problems, where higher values are better. The $x$-axis represents different algorithms, and the $y$-axis represents the AOCC value. The algorithms in Figure~\ref{fig:AOCC_meta_surface_surrogate} run on the surrogate model of \textit{meta-surface}, while the algorithms in Figure~\ref{fig:AOCC_meta_surface_solver} run on the real solver of \textit{meta-surface}.}
    \label{fig:AOCC}
\end{figure*}

\subsection{Experimental Configuration}

To evaluate these hypotheses, we compare LLaMEA's performance across three distinct environments:
\begin{itemize}
    \item Real-World Direct: LLaMEA optimizes directly on real-world problem instances.
    \item Proxy-Driven: LLaMEA optimizes on the top 3 proxy functions identified by our GP generator as most similar to the target real-world landscape.
    \item Benchmark-Driven: LLaMEA optimizes on the top 3 most similar instances from the BBOB test suite (24 standard functions). 
\end{itemize}
To study the importance of landscape fidelity in proxy-driven algorithm discovery, we include a mismatched proxy condition based on standard BBOB benchmark functions. While widely used for algorithm development, these instances are not constructed to match the landscape characteristics of the real-world problems considered here. They therefore serve as low-similarity proxy landscapes in contrast to our ELA-guided GP-generated proxies.

For each experimental configuration, the candidate algorithm that achieves the highest Area Over Convergence Curve (AOCC) — a performance metric commonly used in previous LLaMEA studies~\cite{van2024llamea,yin2025optimizing} — is selected from the pool of 500 generated algorithms for final evaluation. Subsequently, these champion algorithms are compared against established baselines, including Random Search (RS), DE and LSHADE~\cite{rastrigin1963convergence,storn1997differential,aranguren2021improving}. To maintain experimental consistency, the DE and LSHADE baselines are implemented using the Modular DE framework~\cite{vermetten2023modular}, with hyperparameter configurations aligned with established benchmarks in the prior literature~\cite{vermetten2024large,vermetten2025ma}. To account for stochastic variability, results are reported over 10 independent trials. Each execution is restricted to a stringent budget of $50 \times D$ evaluations, representing a highly resource-constrained optimization scenario.

\subsection{Implementation Details}
The ELA features serve as the bridge connecting real-world problems to proxy functions. For each problem of dimension $D$, we sample points $N_{sample}=coef_{ELA}\times D=150\times D$. This coefficient ensures sufficiently dense sampling across different dimensions to capture landscape characteristics. To mitigate fluctuations from single-sample randomness, we randomly select 80\% (i.e., $rate_{ELA}=0.8$) of the total sampling points to compute the ELA feature vector, repeating this process $n_{ELA}=5$ times. And then remove the features that are highly correlated with other remaining features, which have higher correlation than $threshold_{corr}=0.9$. 

For the proxy function generator, we set $n_{pop}=50$ and $n_{gen}=50$. This setting balances search breadth with computational cost, permitting sufficient generations for optimization without becoming excessively time-consuming.
A crossover probability of $p_c=0.5$ and a subtree mutation probability of $p_m=0.1$ represent common GP settings, designed to maintain a balance between population diversity and convergence. The minimum depth of individual trees is set to $min_{depth}=3$, with a maximum depth of $max_{depth}=12$. This permits the generation of sufficiently expressive complex functions while avoiding the creation of excessively large trees that would be difficult to interpret or evaluate through the depth constraint.

We employ the LLaMEA framework as the engine for algorithm generation, integrating it into our framework and utilizing \texttt{gpt-4o-2024-05-13} as the main LLM~\cite{van2024llamea}. Its robust code generation and reasoning capabilities have proven suitable for the synthesis task of meta-heuristic algorithms. Each run of LLaMEA to generate and iterate algorithms is capped at 100 iterations, and for each iteration, only the $50\times D$ evaluation budget is allowed for the algorithm generated to find the best solution. AOCC is selected as feedback for LLaMEA, which has been shown to be efficient in guiding the discovery of the algorithm~\cite{van2024llamea}. Internally, LLaMEA employs a \texttt{(1+1)} ES to iteratively refine the algorithm generated. This search strategy has been extensively validated through experimentation. All experimental data are publicly available~\cite{anonymous_2026_18385405}.



\begin{figure*}[!tb]
    \centering
    \begin{subfigure}[b]{\columnwidth}
        \centering
        \includegraphics[width=\textwidth]{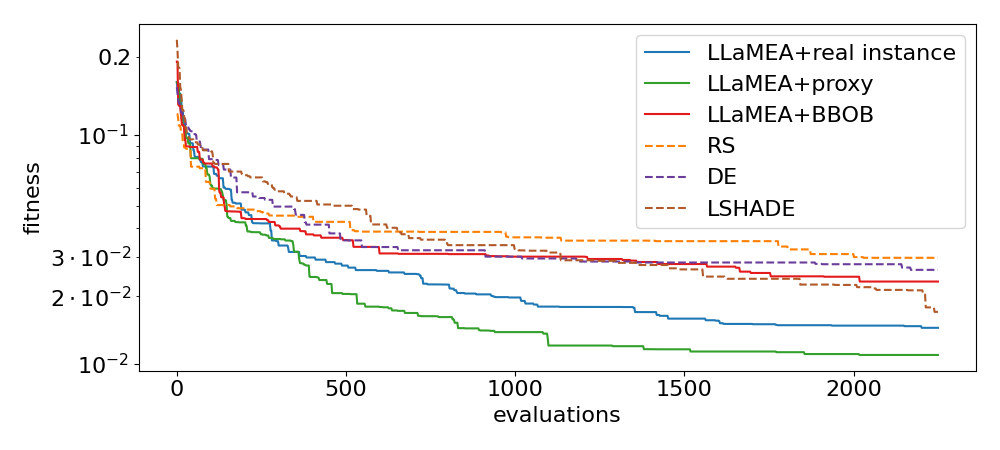}
        \caption{\textit{meta-surface} (surrogate model)}
        \label{fig:curve_meta_surface_surrogate}
    \end{subfigure}
    \begin{subfigure}[b]{\columnwidth}
        \centering
        \includegraphics[width=\textwidth]{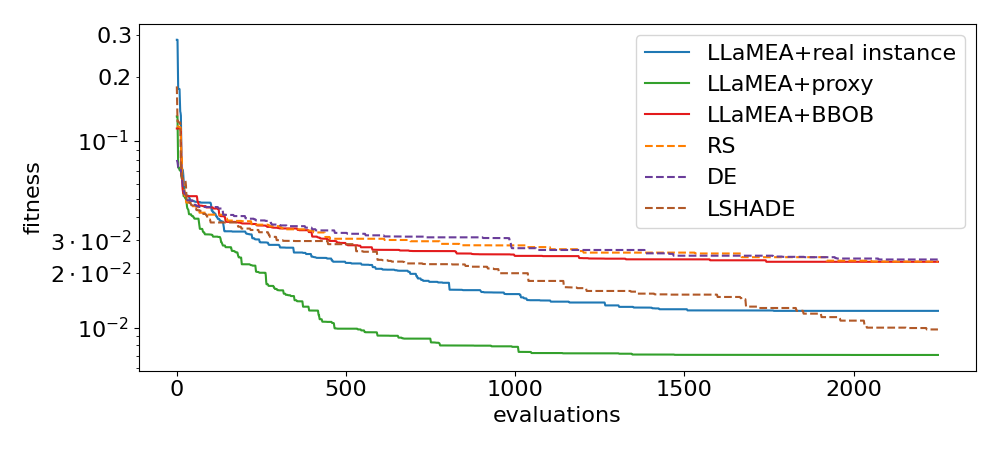}
        \caption{\textit{meta-surface} (real solver)}
        \label{fig:curve_meta_surface_solver}
    \end{subfigure}
    \begin{subfigure}[b]{\columnwidth}
        \centering
        \includegraphics[width=\textwidth]{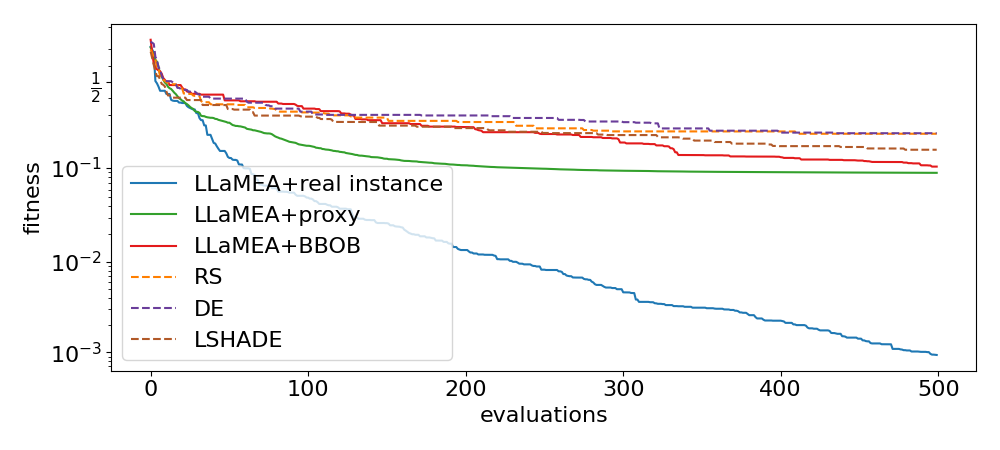}
        \caption{\textit{mini-Bragg}}
        \label{fig:curve_10bragg}
    \end{subfigure}
    \begin{subfigure}[b]{\columnwidth}
        \centering
        \includegraphics[width=\textwidth]{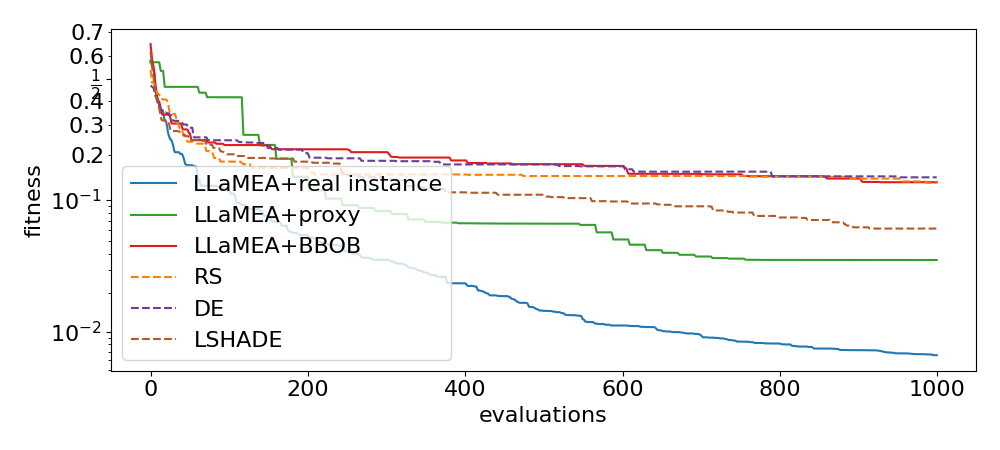}
        \caption{\textit{Bragg}}
        \label{fig:curve_20bragg}
    \end{subfigure}
    \begin{subfigure}[b]{\columnwidth}
        \centering
        \includegraphics[width=\textwidth]{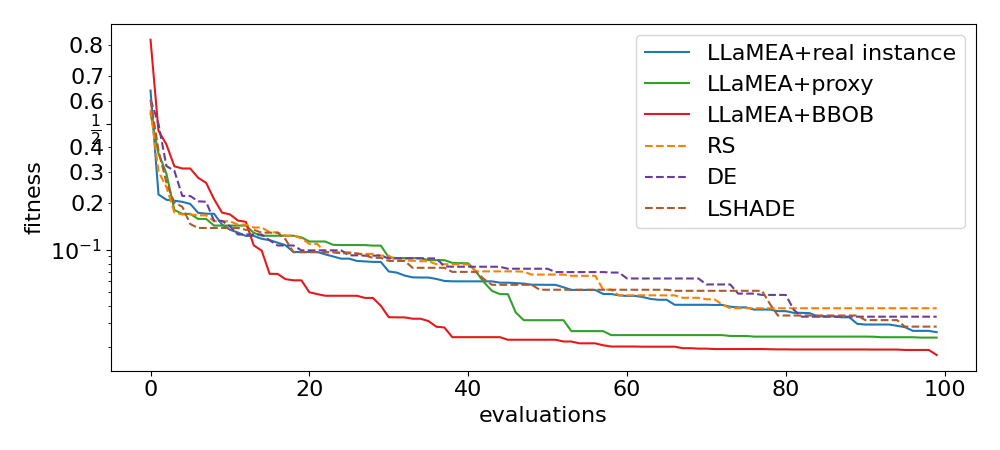}
        \caption{\textit{ellipsometry}}
        \label{fig:curve_ellipsometry}
    \end{subfigure}
    \begin{subfigure}[b]{\columnwidth}
        \centering
        \includegraphics[width=\textwidth]{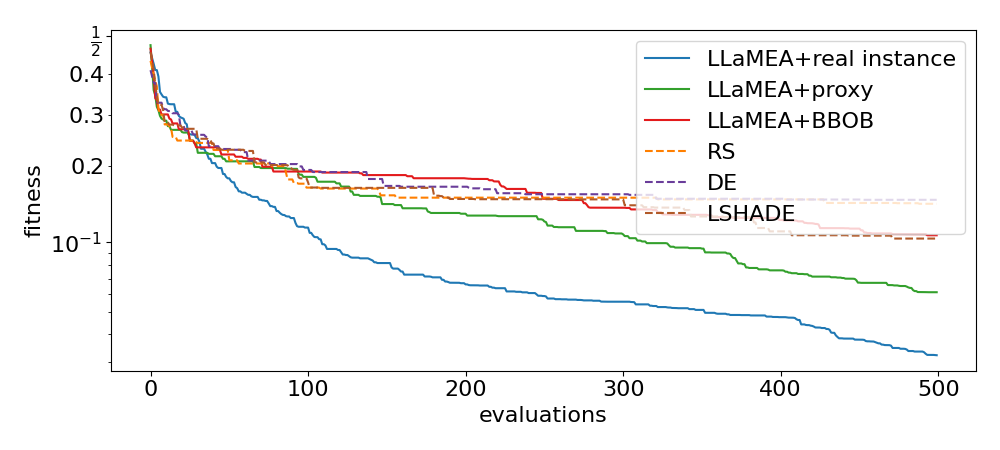}
        \caption{\textit{photovoltaic}}
        \label{fig:curve_photovoltaic}
    \end{subfigure}
    \caption{Benchmark results of the generated and baseline algorithms. The $x$-axis represents the evaluations of the problem and the $y$-axis represents the normalized fitness value, which are all smaller, the better. Each curve is averaged over 10 runs. For most real-world problems, algorithms generated with proxy functions demonstrate performance similar to algorithms derived from real-world problems while simultaneously exhibiting marked superiority over algorithms based on artificially designed problem discovery.}
    \label{fig:curves}
\end{figure*}

\section{Results and Discussions}
\label{sec:results}
This section presents the experimental results of the proposed landscape-aware AAD framework for multiple real-world single-objective continuous optimization problems. We focus on evaluating whether algorithms discovered on proxy functions can be effectively transferred to the original real-world problems under a tight evaluation budget. 

\begin{table}[!tb]
\centering
\caption{Average Wasserstein distance: real-world instances vs. top-3 proxy functions and vs. top-3 BBOB instances. Compared to BBOB instances, proxy functions share much similar landscape characteristics with real-world instances.}
\label{tab:wasserstein_distance}
\resizebox{\columnwidth}{!}{%
\begin{tabular}{@{}cccccc@{}}
\toprule
 & \textit{meta-surface} & \textit{mini-Bragg} & \textit{Bragg} & \textit{ellipsometry} & \textit{photovoltaic} \\ \midrule
proxy     & 4.46         & 2.21       & 3.04  & 2.16         & 0.99         \\
BBOB      & 50.81        & 24.46      & 81.85 & 705.67       & 17.77        \\ \bottomrule
\end{tabular}%
}
\end{table}

Table~\ref{tab:wasserstein_distance} shows the average Wasserstein distance between the ELA features of the real-world instances and those of their corresponding top‑3 most similar proxy functions, as well as the top‑3 most similar BBOB instances. The results clearly show that the proxy functions achieve smaller Wasserstein distances to real‐world problems than the BBOB instances. This structural closeness confirms that our proxy generation successfully captures key landscape features of the target problems, whereas the BBOB suite — included here as an ablative reference — exhibits substantially lower similarity.

Figure~\ref{fig:AOCC} presents the violin plots of the AOCC distributions for three baseline methods alongside three LLaMEA-generated algorithms under varying configurations. In this context, higher AOCC values signify superior optimization performance. Across the majority of the five real-world optimization problems, algorithms evolved via proxy functions consistently outperformed RS, DE, LSHADE and BBOB-derived variants, aligning with our primary objectives, with the \textit{ellipsometry} real-world problem being the sole outlier. The consistently inferior performance of BBOB-driven discovery compared to ELA-matched proxies highlights that algorithm transfer critically depends on landscape similarity rather than merely the availability of cheap proxy problems. Furthermore, the ``direct discovery'' method generally yielded the highest efficacy, except in the case of the \textit{meta-surface} problem. A detailed examination of the 20-layer Bragg reflector (\textit{Bragg}, Figure~\ref{fig:AOCC_20bragg}) reveals that the proxy-driven algorithm achieved substantially higher AOCC values than RS, DE, and LSHADE, indicating better convergence. These results underscore the framework's capacity to autonomously design novel optimization strategies that match or exceed the efficiency of established metaheuristics.

However, a few exceptions are also interesting to discuss. As illustrated in Figure~\ref{fig:AOCC_ellipsometry}, the AOCC of the proxy-driven algorithm trails behind the BBOB-driven variant in the \textit{ellipsometry} instance. Given that AAD is inherently a stochastic process, and LLM further amplifies this stochasticity, the occasional suboptimal performance observed across all instances remains consistent with our expectations. Moreover, as \textit{ellipsometry} is a low-dimensional problem, the proxy functions might prioritize complex landscape features that are less relevant here, suggesting that the framework's relative advantage in low-dimensional spaces requires more targeted investigation. Intriguingly, the proxy-driven method outperformed direct discovery on the \textit{meta-surface} problem. This phenomenon suggests that proxy functions may serve as an effective ``information filter.'' While direct discovery is exposed to the raw, often noisy feedback of real-world instances, the proxy-driven approach utilizes a distilled abstraction of the problem space. By filtering out empirical noise, this method prevents the LLM from over-specializing on specific problem artifacts, thereby yielding a more robust and generalized optimization strategy for complex tasks like the \textit{meta-surface} design.

Figure \ref{fig:curves} illustrates the optimization trajectory by plotting the best-attained fitness against total evaluations. While AOCC serves as a useful summary, these convergence curves, averaged over ten independent runs, provide a more granular perspective on the optimization dynamics and the efficiency of the generated algorithms. Firstly, the current evaluation limit of $50\times D$ appears to be a restrictive threshold which many generated algorithms maintain a downward trajectory at the cut-off point, suggesting potential for further improvement with extended budgets. Notably, proxy-driven algorithms consistently outperform BBOB-driven alternatives, delivering faster convergence and higher-quality final solutions. Within the \textit{meta-surface} and \textit{Bragg} instances, the proxy approach provides a "rapid-start" benefit, achieving major fitness reductions well ahead of other baselines. The strong correlation between Figures~\ref{fig:curves}(a) and~\ref{fig:curves}(b) reinforces the theory that proxy functions act as robust filters; consequently, these algorithms transition seamlessly from clean surrogate environments to noisy, real-world applications without a loss in efficiency.

\begin{figure}[!tb]
    \centering
    \begin{subfigure}[b]{0.9\columnwidth}
        \centering
        \includegraphics[width=\textwidth]{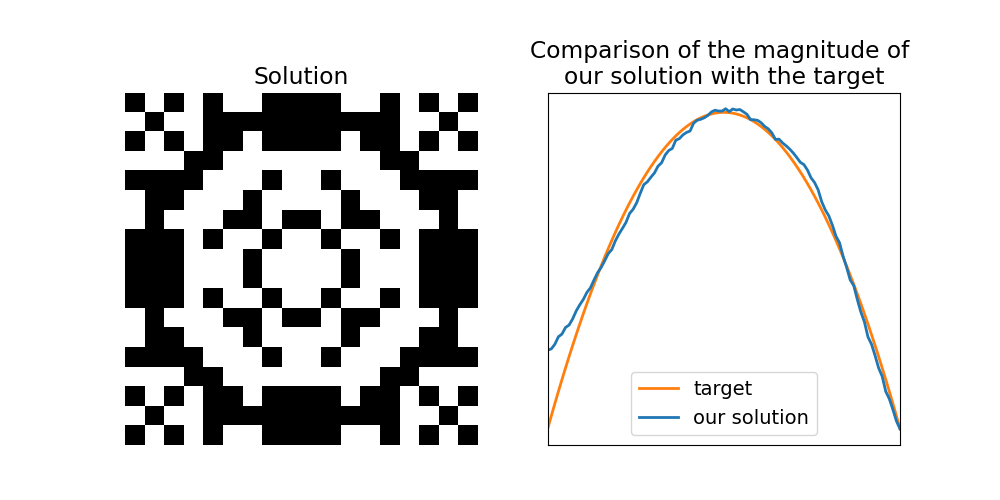}
        \caption{\textit{meta-surface}}
        \label{fig:meta_surface_solution}
    \end{subfigure}
    \centering
    \begin{subfigure}[b]{0.9\columnwidth}
        \centering
        \includegraphics[width=\textwidth]{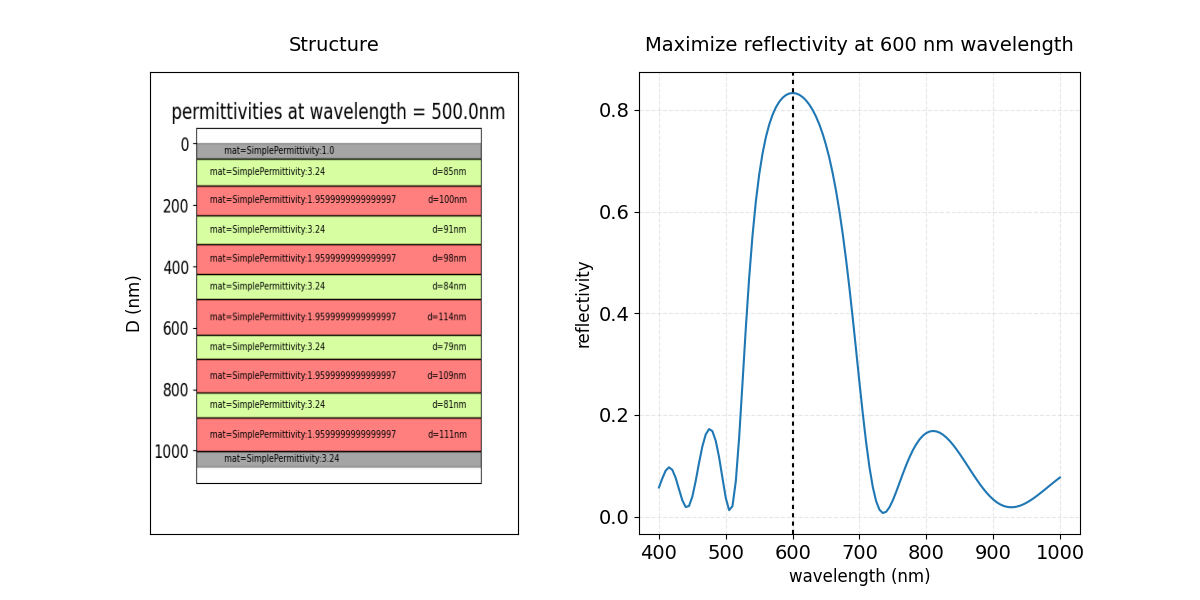}
        \caption{\textit{mini-Bragg}}
        \label{fig:Bragg_solution_10}
    \end{subfigure}
    \centering
    \begin{subfigure}[b]{0.9\columnwidth}
        \centering
        \includegraphics[width=\textwidth]{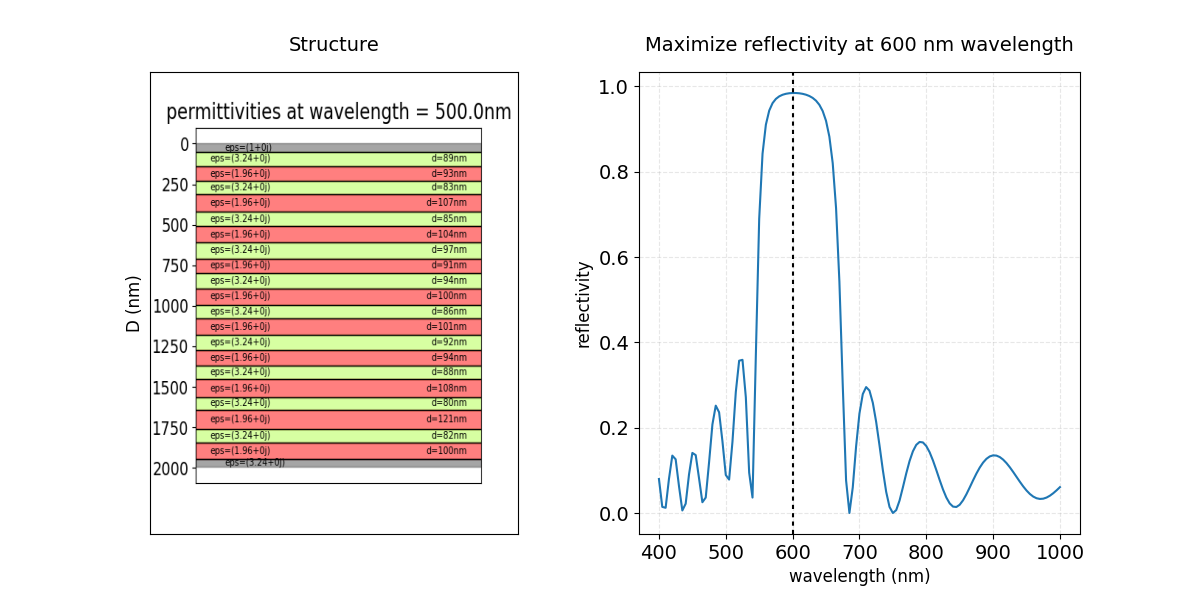}
        \caption{\textit{Bragg}}
        \label{fig:Bragg_solution_20}
    \end{subfigure}
    \centering
    \begin{subfigure}[b]{0.9\columnwidth}
        \centering
        \includegraphics[width=\textwidth]{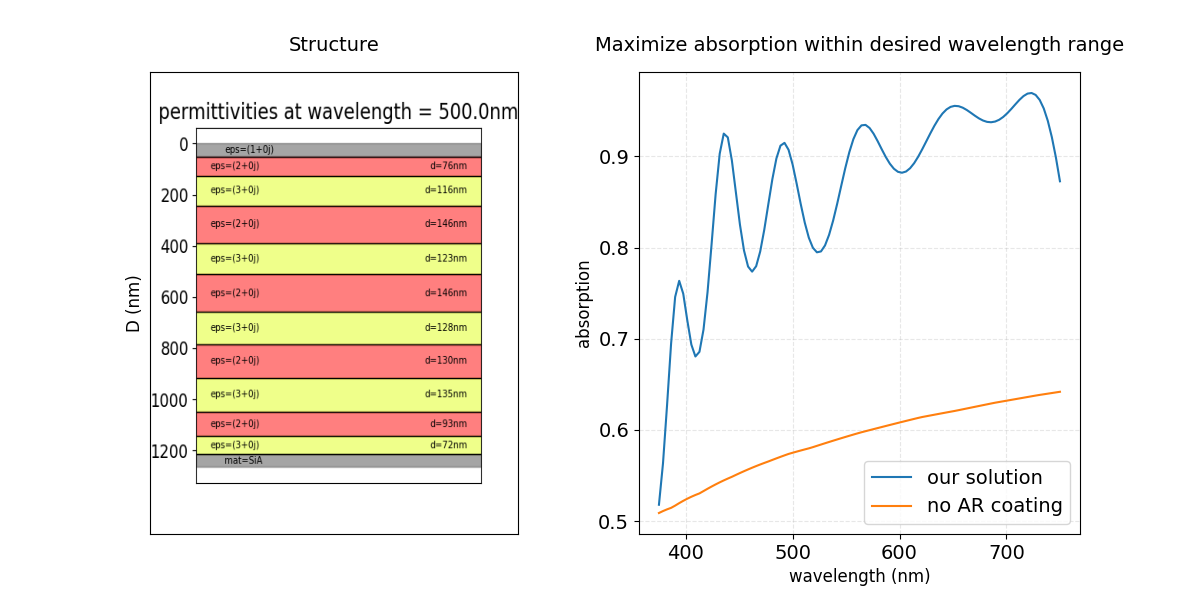}
        \caption{\textit{photovoltaic}}
        \label{fig:photovoltaic_solution}
    \end{subfigure}
\caption{Best solutions found by proxy-driven algorithms.}
\label{fig:solution}
\end{figure}

Figure~\ref{fig:solution} displays the optimal solutions found by proxy-driven algorithms on multiple real-world problems. As shown, the resulting solutions satisfy all physical and engineering constraints while achieving or approaching theoretical optimal values for optical performance metrics such as reflectance and absorptance. For example, in Figure~\ref{fig:meta_surface_solution}, the resulting microstructure patterns precisely control the phase and amplitude to perform the target profile. In optimization of the photovoltaic anti-reflective coating, as shown in Figure~\ref{fig:photovoltaic_solution}, the solution enhances the efficiency of light absorption within the target spectral band. These results not only validate the effectiveness of proxy functions in capturing problems' landscape characteristics but also demonstrate that proxy-driven discovery algorithms can generate reasonable, stable, and high-performance solutions for real-world problems. Consequently, they offer high reliability and practical value in real-world applications.

\section{Conclusions and Future Work}
\label{sec:conclusions}
This paper proposes a landscape-aware automated algorithm design framework that streamlines how LLMs discover optimization algorithms for expensive real-world challenges. By decoupling discovery from high-cost evaluations, it employs a Genetic Programming proxy generator to replicate the target's landscape. By minimizing the Wasserstein distance between Exploratory Landscape Analysis features, the system generates low-cost proxies that preserve the essential structural characteristics of real-world problems. Empirical validation in photonics and optical engineering confirms that algorithms evolved via these GP proxies transfer seamlessly to complex, real-world tasks, consistently outperforming those derived from conventional manual benchmarks. Even under stringent evaluation constraints, this framework produces optimization strategies that rival or exceed established baselines such as DE and LSHADE. These findings validate the applicability of our framework for computationally intensive real-world optimization problems, where direct evaluation-driven discovery is often impractical.


Future efforts will focus on enriching landscape characterization by integrating advanced ELA features and deep-learning-based representations~\cite{van2023doe2vec,seiler2025deep}. By drawing on Neural Architecture Search (NAS), we aim to formulate proxy generation as a differentiable task, enabling the creation of complex, non-linear proxies. On the algorithmic discovery front, we plan to transition toward multi-agent LLM architectures to facilitate iterative logic calibration and real-time error correction. Finally, rigorous interpretability studies of the evolved heuristics will be conducted to map the theoretical synergy between problem landscapes and algorithmic efficiency, ensuring the framework is as transparent as it is high-performing.

\bibliographystyle{ACM-Reference-Format}
\bibliography{sample-base}

\end{document}